\documentclass[fontsize=9pt,a4paper,twocolumn,hidelinks]{scrartcl}

\usepackage[utf8]{inputenc}

\usepackage[english]{babel}

\usepackage{glossaries-extra}
\glsdisablehyper

\newabbreviation{bss2}{BSS-2}{BrainScaleS-2}
\newabbreviation{snpe}{SNPE}{sequential neural posterior estimation}
\newabbreviation{psp}{PSP}{post-synaptic potential}
\newabbreviation{lfi}{LFI}{likelihood-free inference}
\newabbreviation{sbi}{SBI}{simulation-based inference}
\newabbreviation{ppc}{PPC}{posterior-predictive check}
\newabbreviation{maf}{MAF}{masked autoregressive flow}
\newabbreviation{adex}{AdEx}{adaptive exponential integrate-and-fire}
\newabbreviation{lif}{LIF}{leaky integrate-and-fire}
\newabbreviation{nde}{NDE}{neural density estimator}
 \usepackage{graphicx}
\graphicspath{{./fig/}}

\usepackage{tikz}
\usetikzlibrary{positioning,shapes.multipart, math, shadows.blur, external}
\usepackage{circuitikz}
\tikzset{
	panel/.style={inner sep=0, outer sep=0, anchor=north west},
	fig label/.style={label, outer sep=0, inner sep=1pt, anchor=north west, font=\small\bfseries},
}
\usepackage{physics}
\usepackage{hyperref}
\usepackage{csquotes}
\usepackage{neuralnetwork}
\usepackage{circledsteps}  \usepackage{bm} 

\usepackage{pgfplots}

\usepackage{authblk}

\usepackage[binary-units=true]{siunitx}

\usepackage[noabbrev,capitalise]{cleveref}

\usepackage{changes}
\usepackage{todonotes}

\newlength{\singlecolumn}
\newlength{\onehalfcolumn}
\newlength{\doublecolumn}
\definecolor{comp0}{HTML}{66A61E}
\definecolor{comp1}{HTML}{E6AB02}
\definecolor{comp2}{HTML}{A6761D}
\definecolor{comp3}{HTML}{666666}

\newcommand{\formatpanel}[1]{(\lowercase{#1})}
\newcommand{\subfig}[2]{\cref{#1}\formatpanel{#2}}

\newcommand{\myvec}[1]{\bm{#1}}
\newcommand{\mymat}[1]{\bm{#1}}

\newcommand{\subcaption}[1]{\textbf{\formatpanel{#1}}}

\newcommand{\figBSS}{\subfig{fig:bsssbi}{A}}
\newcommand{\figSbi}{\subfig{fig:bsssbi}{B}}

\newcommand{\figEval}{\subfig{fig:evaluation}{A}}
\newcommand{\figGridsearch}{\subfig{fig:evaluation}{B}}
\newcommand{\figGridsearchDiff}{\subfig{fig:2d}{A}}
\newcommand{\figTracesRelative}{\subfig{fig:2d}{B}}
\newcommand{\figPosterior}{\subfig{fig:2d}{C}}
\newcommand{\figSamples}{\subfig{fig:2d}{D}}
\newcommand{\figTracesAbsolute}{\subfig{fig:2d}{E}}

\newcommand{\figMarginals}{\subfig{fig:md}{A}}
\newcommand{\figCorrelations}{\subfig{fig:md}{B}}
\newcommand{\figPPC}{\subfig{fig:md_validation}{A}}
\newcommand{\figCoverageMd}{\subfig{fig:md_validation}{B}}

\newcommand{\figPosteriorEvo}{\subfig{fig:simulations}{A}}

\title{Simulation-based Inference for Model Parameterization on Analog Neuromorphic Hardware}

\author[a,2]{Jakob Kaiser}
\author[a]{Raphael Stock}
\author[a]{Eric M\"uller}
\author[a,2]{Johannes Schemmel}
\author[b]{Sebastian Schmitt}

\affil[a]{Kirchhoff-Institute for Physics (European Institute for Neuromorphic Computing), Heidelberg University, Heidelberg, Germany}
\affil[b]{Department for Neuro- and Sensory Physiology, University Medical Center G{\"o}ttingen, G{\"o}ttingen, Germany}
 
\usepackage[style=authoryear,natbib=true,maxcitenames=2,uniquelist=false]{biblatex}
\affil[2]{Corresponding authors (\href{mailto:jakob.kaiser@kip.uni-heidelberg.de}{jakob.kaiser@kip.uni-heidelberg.de} or \href{mailto:schemmel@kip.uni-heidelberg.de}{schemmel@kip.uni-heidelberg.de})}

\usepackage[format=plain]{caption}
\usepackage{tcolorbox}
\usepackage{cuted}  

\begin{document}
	\begin{strip}
		\begin{tcolorbox}[colback=comp0!30,colframe=comp0!50]
This is the Accepted Manuscript version of an article accepted for publication in \emph{Neuromorphic Computing and Engineering}.
IOP Publishing Ltd is not responsible for any errors or omissions in this version of the manuscript or any version derived from it.
This Accepted Manuscript is published under a~CC BY license.
The Version of Record is available online at \url{https://doi.org/10.1088/2634-4386/ad046d}.
\end{tcolorbox}

 		\maketitle
		\begin{abstract}
			\bfseries
			The \gls{bss2} system implements physical models of neurons as well as synapses and aims for an energy-efficient and fast emulation of biological neurons.
When replicating neuroscientific experiments on \gls{bss2}, a major challenge is finding suitable model parameters.
This study investigates the suitability of the \gls{snpe} algorithm for parameterizing a multi-compartmental neuron model emulated on the \gls{bss2} analog neuromorphic system.
The \gls{snpe} algorithm belongs to the class of \gls{sbi} methods and estimates the posterior distribution of the model parameters; access to the posterior allows quantifying the confidence in parameter estimations and unveiling correlation between model parameters.

For our multi-compartmental model, we show that the approximated posterior agrees with experimental observations and that the identified correlation between parameters fits theoretical expectations.
Furthermore, as already shown for software simulations, the algorithm can deal with high-dimensional observations and parameter spaces when the data is generated by emulations on \gls{bss2}.

These results suggest that the \gls{snpe} algorithm is a promising approach for automating the parameterization and the analyzation of complex models, especially when dealing with characteristic properties of analog neuromorphic substrates, such as trial-to-trial variations or limited parameter ranges.
 		\end{abstract}

		Keywords: analog $|$ neuromorphic $|$ simulation-based inference $|$ multi-compartment
 	\end{strip}

	\section{Introduction}\label{sec:introduction}
\glsresetall

Mechanistic models, which try to explain the causality between inputs and outputs, are integral to scientific research.
On the one hand they can increase the understanding of the mechanisms which cause the phenomena and on the other make predictions about new outcomes which can then be tested experimentally \citep{baker2018mechanistic}.
After a mechanistic model has been formulated, one of the remaining challenges is to find suitable model parameters which lead to a close agreement between model behavior and experimental observations.

Several approaches such as the hand-tuning of parameters, grid searches, random/stochastic searches, evolutionary algorithms, simulated annealing and particle swarm algorithms have been used in neuroscience to find appropriate model parameters \citep{vanier1999comparative, vangeit2008automated}.
Drawbacks of these methods are that they rely on a score which represents how close the results of a simulated model are to the target observation and that they in general only yield the best performing set of parameters.
Furthermore, these algorithms are often computationally expensive since they require many simulations to find suitable parameters \citep{goncalves2020training}.

The class of \gls{sbi} algorithms makes statistical inference methods available for models where the likelihood is not tractable and provides an approximation of the posterior distribution of the model parameters.
Advantages of deriving an approximation of the posterior include the possibility to find correlations between model parameters and to evaluate the confidence in the estimated parameters.
Early \gls{sbi} approaches  rely on defining a score and are computationally inefficient since they disregard many simulation which have a low score \citep{sisson2018handbook}.

Recent advances in machine learning lead to a new class of \gls{sbi} algorithms which promise to be computationally more efficient and do not depend on a score function \citep{papamakarios2016fast, lueckmann2017flexible, greenberg2019automatic, cranmer2020frontier, deistler2022truncated}.
In this paper we will focus on the \gls{snpe} algorithm which was already applied to infer parameters for different neuroscientific models \citep{lueckmann2017flexible, goncalves2020training}.
More specifically, we want to investigate if this algorithm is suitable to parameterize neuron models which are emulated on the \gls{bss2} analog neuromorphic hardware system \citep{pehle2022brainscales2_nopreprint_nourl}.

Neuromorphic computation draws inspiration from the brain to find time and energy efficient computing architectures as well as algorithms \citep{indiveri2011neuromorphic}.
The \gls{bss2} system emulates the behavior of neurons and synapses on analog circuits in continuous time \citep{billaudelle2022accurate} and does not solve the model equations mathematically like digital neuromorphic hardware \citep{furber2012overview, davies2018loihi, mayr2019spinnaker}.

In previous experiments on the \gls{bss2} system, hardware parameters were set by calibration routines, grid searches, gradient-based optimization or by hand-tuning \citep{billaudelle2022accurate, cramer2022surrogate, pehle2023event, aamir2018dls3neuron, wunderlich2019advantages, kaiser2022emulating}.
The hand-tuning of parameters can be tedious and relies on the domain-specific knowledge of the experimenter such that automated parameter-search methods are inevitable for complex problems \citep{vanier1999comparative}.
Similarly, a calibration routine can only be formulated if the relationship between parameters and observations is known.
Depending on the dimensionality of the parameter space, grid searches and random searches can be computationally too expensive.
The \gls{snpe} algorithm promises to find approximations of the posterior even if the parameter space is high-dimensional and the relationship between the parameters and the observation is unknown \citep{lueckmann2017flexible, greenberg2019automatic, goncalves2020training}.

Furthermore, the \gls{snpe} algorithm is designed for probabilistic models.
This makes it a suitable choice for models which deal with intrinsic probabilistic behavior such as analog neuromorphic hardware which is subject to temporal noise.

In the present study we emulated a passive multi-compartmental neuron model on \gls{bss2} and investigated whether the \gls{snpe} algorithm can find suitable model parameters to reproduce previously recorded target observations.
For a two-dimensional parameter space, we show that the approximated posterior derived with the \gls{snpe} algorithm agreeed with a grid search over the whole parameter space and that the correlations between model parameters are in agreement with theoretical predictions.

Finally, we extended the problem to a higher-dimensional~(7) parameter space and examined the approximated posteriors with \glspl{ppc}.
The correlations between parameters of this high-dimensional model did agree with the model equations.

All in all, our results indicate that the \gls{snpe} algorithm is able to deal with the intrinsic trial-to-trial variations of analog neuromorphic hardware and is able to approximate posterior distributions which are in agreement with the given target observations.
 	\section{Methods}\label{sec:methods}
This section starts by introducing the \gls{bss2} neuromorphic system.
We chose the attenuation of \glspl{psp} in a passive chain of compartments to test if the \gls{snpe} algorithm is capable to parameterize experiments on \gls{bss2}.
Therefore, we introduce the attenuation experiment before we describe the \gls{snpe} algorithm.
We conclude this section by introducing methods which we used to validate our posterior approximations.

\begin{figure*}
   \includegraphics{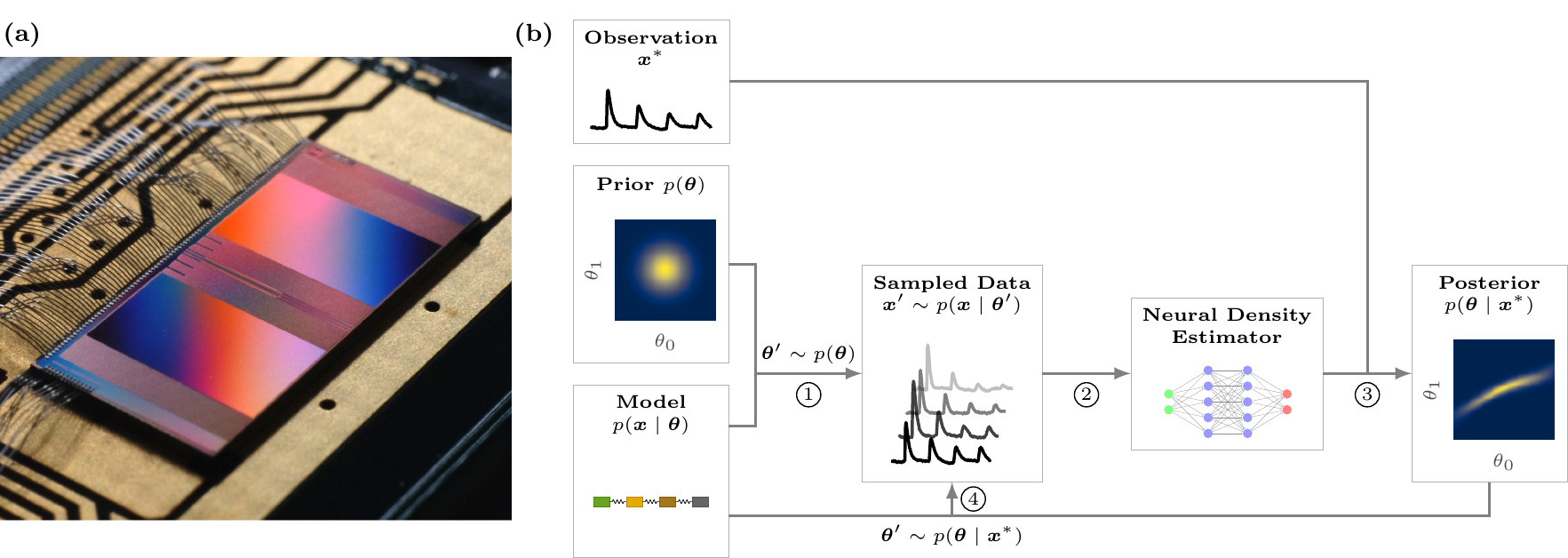}
	\caption{
   	The \glsfmtfirst{bss2} system and the \glsfmtfirst{snpe} algorithm.
   	\subcaption{A} Photograph of the \glsfmttext{bss2} neuromorphic chip bonded to a carrier board.
   	\subcaption{B} Visualization of the \glsfmttext{snpe} algorithm \citep{papamakarios2016fast, lueckmann2017flexible, greenberg2019automatic}.
			This algorithm can be used to find an approximation for the posterior distribution $p(\myvec{\theta} \mid \myvec{x}^*)$ of parameters $\myvec{\theta}$ which recreate a target observation $\myvec{x}^*$.
   		The target observation $\myvec{x}^*$, a prior belief about the parameter distribution $p(\myvec{\theta})$ and a model which gives implicit access to the likelihood $p(\myvec{x} \mid \myvec{\theta})$ are given as inputs to the algorithm.
   		In step \Circled{1}, we sample parameters $\myvec{\theta}'$ from the prior distribution and the model is evaluated with these parameters to obtain observations $\myvec{x}'$.
   		This implicitly allows us to sample form the likelihood $p(\myvec{x} \mid \myvec{\theta}')$.
		In the following step \Circled{2}, the set of parameters and the corresponding observations are used to train a \glsfmtfirst{nde}.
		The \glsfmttext{nde} serves as a surrogate for the posterior distribution $p(\myvec{\theta} \mid \myvec{x})$.
			Frequently, we are interested in a single observation $\myvec{x}^*$ and we can restrict the \glsfmttext{nde} to this observation, step \Circled{3}.
			We can now use samples drawn from the posterior $\myvec{\theta}' \sim p(\myvec{\theta} \mid \myvec{x}^*)$ to generate new samples and retrain the \glsfmttext{nde}, repeating step \Circled{2} and \Circled{3}.
			Steps \Circled{2} to \Circled{4} can be repeated several times to improve the estimate of the posterior.
			The figure is based on \citep[Figure~1]{goncalves2020training}.}
	\label{fig:bsssbi}
\end{figure*}

\subsection{\Glsfmtlong{bss2}}\label{sec:bss}
\Gls{bss2} is a mixed-signal analog neuromorphic system; neuron and synapse dynamics are emulated by analog circuits while spike events and configuration data rely on digital communication, \figBSS.
More specifically, the dynamics of the analog neuron circuits are designed to resemble the dynamics of the \gls{adex} neuron model \citep{brette2005adaptive, billaudelle2022accurate}.
Voltages and currents on these analog circuits directly represent the state of the emulated neuron.

\subsubsection{Neuron Dynamics}
The \gls{adex} neuron model extends the \gls{lif} neuron model by introducing an exponential and an adaptation current \citep{brette2005adaptive}.
The high configurability, see below, of the \gls{bss2} system allows disabling these currents to model \gls{lif} neurons.
Furthermore, several neuron circuits can be connected to form multi-compartmental neuron models \citep{kaiser2022emulating}.

In this publication, we will consider multi-compartmental neuron models for which the membrane potentials in the different compartments $V_\text{m}$ adhere to the dynamics of the \gls{lif} neuron model,
\begin{equation}\label{eq:neuron}
	C_\text{m} \dv{V_\text{m}(t)}{t} = g_\text{leak} \cdot \left( V_\text{leak} - V_\text{m}(t) \right)
									 + I_\text{syn}(t) + I_\text{axial}(t),
\end{equation}
where $C_\text{m}$ is the membrane capacitance, $g_\text{leak}$ the leak conductance and $V_\text{leak}$ the leak potential.

The two currents in \cref{eq:neuron} arise due to synaptic input, $I_\text{syn}$, and connections to neighboring compartments, $I_\text{axial}$.
The synaptic current $I_\text{syn}$ models current-based synapses with an exponential kernel.
The current $I_{\text{axial},i}(t)$ on compartment $i$\footnote{Since all variables in \cref{eq:neuron} refer to compartment $i$, we omitted the subscript $i$ in \cref{eq:neuron} for easier readability.} due to neighboring compartments is given by
\begin{equation}\label{eq:mc_current}
	I_{\text{axial},i}(t) = \sum_j g_\text{axial}^{i \leftrightarrow j} \cdot \left( V_\text{m,j}(t) - V_\text{m,i}(t) \right),
\end{equation}
where the sum runs over all neighboring compartments $\{j\}$, $g_\text{axial}^{i \leftrightarrow j}$ represents the conductance between these compartments and $V_j$ is the membrane potential of the neighboring compartment.

Once the membrane potential $V_\text{m}$ crosses a threshold potential $V_\text{thres}$ a spike is generated and the membrane potential is reset to the reset potential $V_\text{reset}$.\footnote{These digital spikes can be used as inputs to other neurons on the chip or can be recorded as observables.}
After the refractory time $\tau_\text{ref}$ the reset is released and the membrane potential $V_\text{m}$ continues to adhere to the dynamics of \cref{eq:neuron}.

\subsubsection{Configurability}\label{sec:configurability}
The behavior of the neuron circuits on \gls{bss2} can be controlled by several digital and analog parameters.

Digital parameters, for example, control if the adaptation or exponential currents are connected to the membrane \citep{billaudelle2022accurate} and how different neuron circuits are connected to each other to form multi-compartmental neuron models \citep{kaiser2022emulating}.

Analog reference voltages and currents control quantities such as the leak conductance $g_\text{leak}$, leak potential $V_\text{leak}$ or the axial conductance $g_\text{axial}$ between neuron circuits.
These analog references are provided by an analog on-chip memory array which converts digital \SI{10}{\bit} values to currents and voltages \citep{hock13analogmemory}.
Since the last value is reserved, reference currents and voltages can be adjusted digitally from \numrange{0}{1022}.
This large configuration range allows tuning the neuron circuits to a variety of different operating regimes and to compensate manufacturing-induced mismatch between different neuron circuits \citep{billaudelle2022accurate}.

In the current publication, we use the latest revision of the \gls{bss2} system \citep{pehle2022brainscales2_nopreprint_nourl, billaudelle2022accurate}.
The PyNN domain-specific language \citep{davison2009pynn} was used to formulate the experiments and the \gls{bss2} OS to define as well as to control the experiments \citep{mueller2020bss2ll_nourl}.

\subsection{Experiment Description - A Linear Chain of Compartments}

In order to test the capabilities of the \gls{snpe} algorithm, we considered a multi-compartmental model which consisted of a chain of passive compartments, see \figEval{}.
Such multi-compartmental models have been used to model dendrites and axons \citep{rall1962electrophysiology, fatt1951analysis}.
Each compartment $i$ was connected to a leak potential $V_\text{leak}$ via a leak conductance $g_\text{leak}^i$ and to the neighboring compartment via an axial conductance $g_\text{axial}^{i \leftrightarrow i + 1}$, compare \cref{eq:mc_current}.
These conductances served as our parameters $\myvec{\theta}$, all other parameters were fixed.

\begin{figure*}
	\includegraphics{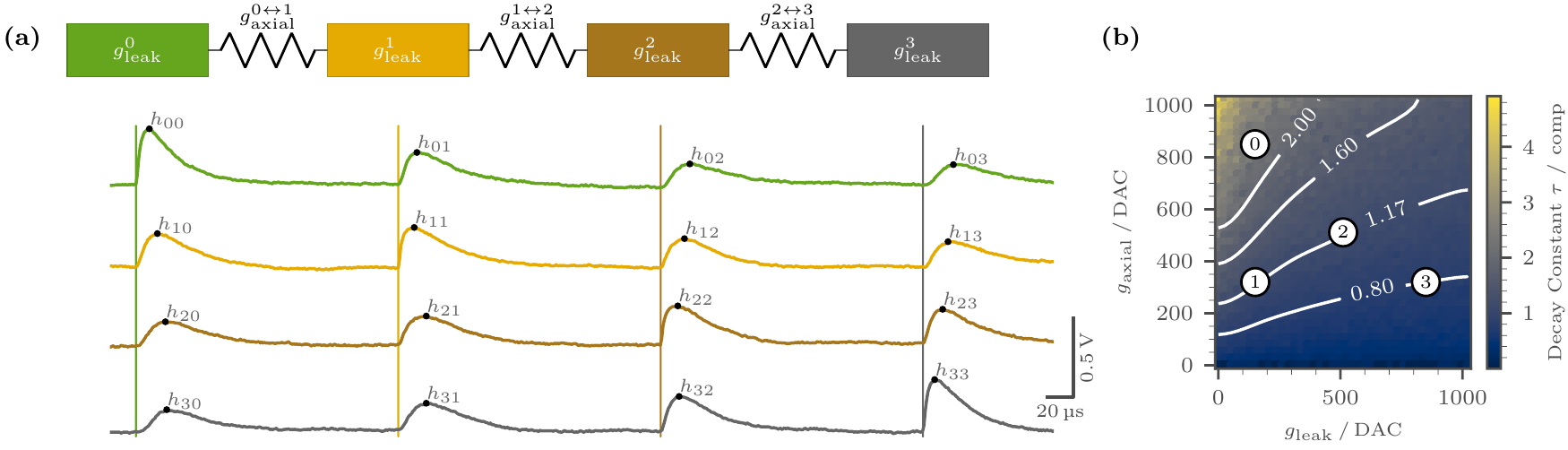}
	\caption{Model of a passive compartment chain and grid search results.
		\subcaption{A} The parameters of the model are given by the leak conductance in each compartment $g_\text{leak}^i$ and the axial conductance between compartments $g_\text{axial}^{i \leftrightarrow i + 1}$.
		In our experiment we observe the propagation of \glsfmtfirstpl{psp}.
		Here we show membrane traces of neurons which were emulated on the \glsfmtlong{bss2} system.
		We inject a synaptic input (vertical lines) in one compartment after another and record the membrane potential in each compartment (different rows).
		From these traces we extract the heights of the \glspl{psp} $h_{ij}$.
		We use the matrix of all heights $\mymat{H}$, the heights resulting from an input to the first compartment $\myvec{F} = [h_{00}, h_{10}, h_{20}, h_{30}]$ or the decay constant $\tau$ from an exponential fit to $\myvec{F}$ as observables.
		The scale bar in the lower right corner indicates the voltage and time in the hardware domain.
		\subcaption{B} Grid search on \glsfmtlong{bss2} of the decay constant $\tau$; the decay constant is given in units of \enquote{compartments} and calculated by fitting an exponential to the \glspl{psp} which result from an input to the first compartment, compare panel \formatpanel{A}.
			We divided the parameter space in an evenly spaced grid with \num{40} values in each dimension, recorded the resulting \glsfmttext{psp} heights in each compartment and extracted the decay constant $\tau$; \cref{fig:exponential} shows the exponential fits for some exemplary measurements.
			The decay constant $\tau$ decreases as the leak conductance $g_\text{leak}$ is increased or the axial conductance $g_\text{axial}$ is reduced.
			The white contour lines mark regions with equal decay constant and show a correlation between leak and axial conductance.
			Traces recorded at the numbered points are displayed in \cref{fig:2d}.
	}
	\label{fig:evaluation}
\end{figure*}

We injected synaptic inputs in the different compartments and observed how the \glspl{psp} propagate along the chain.
More specifically, we looked at the heights of \glspl{psp}; in the following we will use the notation $h_{ij}$ to describe the \gls{psp} height which was observed in compartment $i$ after an input to compartment $j$, \figEval{}.
Since we were only interested in the passive propagation, we disabled the spiking threshold, this is equivalent to $V_\text{thres} \to \infty$.

Due to the low-pass properties of the passive chain, the response in the first compartment broadened and its height decreased as the synaptic input was injected further away from the first compartment, compare first row in \figEval{}.
A similar behavior was visible when we looked at the voltage traces in the second compartment: the \glspl{psp} broadened and flattened for inputs further away from the recording site.
Since we considered a finite chain, we saw that an input at the end of the chain affected the membrane potential more strongly, for example $h_{10} > h_{12}$.

The height of the \glspl{psp} depended on the leak and axial conductance \citep{fatt1951analysis}.
A higher leak or axial conductance resulted in lower \gls{psp} heights at the injection site as less charge can be accumulated on the compartment.
Therefore, the \gls{psp} heights $\mymat{H}$ or quantities derived form them were suitable observations $\myvec{x}$ that could be used to infer parameters $\myvec{\theta}$.
Besides the full matrix of \gls{psp} heights, we used the \gls{psp} heights which resulted from an input to the first compartment $\myvec{F} = [h_{00}, h_{10}, h_{20}, h_{30}]$ and the decay constant $\tau$ from an exponential fit to $\myvec{F}$ as observables.

\subsection{Sequential Neural Posterior Estimation Algorithm}
The \gls{snpe} algorithm \citep{papamakarios2016fast, lueckmann2017flexible, greenberg2019automatic} belongs to the class of \gls{sbi} algorithms and allows finding an approximation of the posterior distribution $p\left( \myvec{\theta} \mid \myvec{x}^* \right)$ in cases where the likelihood $p\left( \myvec{x} \mid \theta \right)$ is intractable.
Here $\myvec{\theta}$ are the parameters of a mechanistic model for which we try to find parameters which reproduce a target observation $\myvec{x}^*$.
The main idea is to evaluate the model for different parameters $\{ \myvec{\theta}_i \}$, extract the observations $\{ \myvec{x}_i \}$ and fit a flexible probability distribution as a posterior to this set of parameters and observations.
As the name suggests the parameters of these probability distributions are determined by neural networks.

The algorithm takes a target observation $\myvec{x}^*$, prior $p(\myvec{\theta})$ and a model for which suitable parameters should be found as an input, \figSbi.
The prior is used to draw random parameters $\myvec{\theta}' \sim p(\myvec{\theta})$.
By executing the model with the given parameters $\myvec{\theta}'$ we implicitly sample from the likelihood $\myvec{x}' \sim p(\myvec{x} \mid \myvec{\theta}')$.
In our case the evaluation of the model is the emulation on the \gls{bss2} system.

In the second step, a \gls{nde} is trained to approximate the posterior distribution $p(\myvec{\theta} \mid \myvec{x})$.
The \gls{nde} is a flexible set of probability distributions which are parameterized by a neural network.
Typical choices are mixture-density networks \citep{bishop1994mixture, papamakarios2016fast, lueckmann2017flexible, greenberg2019automatic} or \glspl{maf} \citep{papamakarios2017masked, papamakarios2019sequential, goncalves2020training, papamakarios2021normalizing}.
The \gls{nde} is commonly trained by minimizing the negative log-likelihood of the previously drawn samples.
Therefore, unlike traditional \gls{sbi} algorithms the \gls{snpe} algorithm does not depend on a user-defined score function.
After successful training, the \gls{nde} approximates the posterior distribution of the parameters for any observation $\myvec{x}$.

If we are only interested in a single target observation $\myvec{x}^*$, we can use the estimated posterior distribution in the following rounds as a proposal prior \citep{papamakarios2016fast, lueckmann2017flexible, greenberg2019automatic}.
While this sequential approach can increase sample efficiency, the obtained approximation of the posterior is no longer amortized, i.e.\ it can only be used to infer parameters for the target observation $\myvec{x}^*$ and not any arbitrary observation $\myvec{x}$.

In our experiments we applied the algorithm presented in \citet{greenberg2019automatic} which is implemented in the \texttt{Python} package \texttt{sbi}\footnote{\url{https://github.com/mackelab/sbi}, we used version \texttt{0.21.0}.} \citep{tejero2020sbi}.
The structure of the \gls{nde} as well as other hyperparameters of the \gls{snpe} algorithm can be found in \cref{sec:meth:snpe}.

\subsection{Validation}\label{sec:validation}
In order to validate the approximated posteriors we used \glspl{ppc} and calculated the expected coverage for each posterior \citep{hermans2022crisis}.

\subsubsection{Predictive Posterior Check}
We performed \glspl{ppc} to check if an approximated posterior $p\left( \myvec{\theta} \mid \myvec{x}^* \right)$ yielded parameters $\myvec{\theta}$ which are in agreement with the original observation $\myvec{x}^*$.
As discussed in \citet{lueckmann2021benchmarking}, \glspl{ppc} do not measure the similarity of the approximated and true posterior and should just be used as a check rather than a metric.
Nevertheless, we found that \gls{ppc} were sensitive enough to highlight posterior approximations which did not agree with our expectation of the posterior based on grid search results.
In the appendix, we illustrate examples of mismatching posteriors, \cref{fig:simulations}, and show how we used \glspl{ppc} to adjust the hyperparameters of the \gls{nde}, \cref{fig:transforms}.

For all \glspl{ppc} we drew \num{1000} random parameters $\{\myvec{\theta}_i\}$ from the approximated posterior $p\left( \myvec{\theta} \mid \myvec{x}^* \right)$, emulated the chain model with these parameters on \gls{bss2} and recorded the observables $\{\myvec{x}_i\}$.
We used the mean Euclidean distance between these observations and the target observation $\myvec{x}^*$ as an indicator for an successful approximation.

\subsubsection{Expected Coverage}
Recent publications indicate that the posteriors approximated with the \gls{snpe} algorithm tend to yield overconfident posterior approximations, i.e.\ the posterior distribution is to narrow \citep{hermans2022crisis, deistler2022truncated}.
To test the confidence of our posteriors we calculated the expected coverage as suggested in \citet{hermans2022crisis}.

We calculated the expected coverage as follows.
First we drew \num{1000} random samples from the prior distribution, $\{ \myvec{\theta}_i^* \} \sim p(\myvec{\theta})$.
We then performed the experiment with these parameters on \gls{bss2} to obtain observations $\{ \myvec{x}_i^* \}$, yielding pairs $\{ (\myvec{\theta}^*, \myvec{x}^*)_i \} \sim p(\myvec{\theta}, \myvec{x})$.
Finally, we calculated the coverage of each pair $(\myvec{\theta}^*, \myvec{x}^*)_i$ and averaged over them to get the expected coverage.

The coverage of a single pair was calculated as follows.
We drew \num{10000} samples for from the amortized posterior $\{ \myvec{\theta}'_j \} \sim p(\myvec{\theta} \mid \myvec{x}_i^*)$ for each pair (i.e.\ we performed the coverage tests with the first round approximations of the posterior and not the final approximations\footnote{As the first round approximation is amortized, we could condition it on arbitrary observations. Approximations in later rounds are restricted to a single target observation.}).
Next, we used the posterior probability of the original parameter $p(\myvec{\theta}^* \mid \myvec{x}_i^*)$ and of the drawn samples $\{p(\myvec{\theta}'_j \mid \myvec{x}_i^*) \}_j$ to estimate the coverage.
 	\section{Results}\label{sec:results}
To simplify the problem, we started by considering a two-dimensional parameter space.
This was achieved by setting the leak and axial conductance globally.
The low dimensionality of the parameter space allowed us to perform a grid search in a reasonable amount of time and to easily visualize the results.
The grid search result can give an intuition about the behavior of the chain and was used as a comparison to the approximated posterior obtained with the \gls{snpe} algorithm.

We also executed the \gls{snpe} for a high-dimensional~(7) parameter space and performed \glspl{ppc}.
For both, the two-dimensional and high-dimensional parameter space we looked at different kind of observations and how these influence the approximated posterior.
Furthermore, we analyzed the correlation for each posterior and performed coverage tests.

\subsection{Two-dimensional Parameter Space}\label{sec:2d}
We reduced the dimensions of the parameter space to two by setting the leak and axial conductance for all compartments and connections to the same digital value\footnote{Due to the production induced mismatch between analog circuits, the same digital values lead to different conductances on the \gls{bss2} system.}; $g_\text{leak}^i = g_\text{leak}\; \forall i \in \{0, 1, 2, 3\}$ and $g_\text{axial}^{i \leftrightarrow i + 1} = g_\text{axial}\; \forall i \in \{0, 1, 2\}$.

\begin{figure*}
	\includegraphics{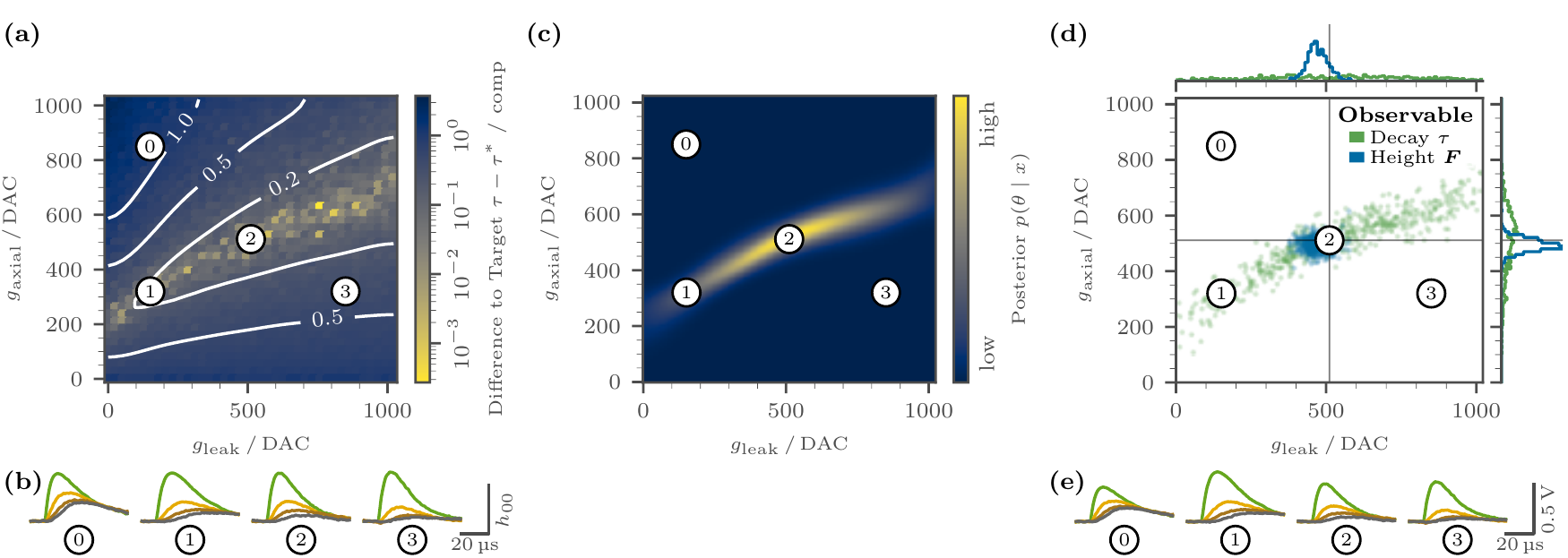}
	\caption{
		Propagation of \glsfmtfirstpl{psp} in a passive chain of four compartments emulated on the \glsfmtlong{bss2} system.
		Leak and axial conductance were set to the same value for all compartments and connections between compartments.
		\subcaption{A} Grid search result illustrated as the difference of the measured decay constant $\tau$, compare \figGridsearch{}, to the target decay constant $\tau^*$: $\left|\tau - \tau^*\right|$.
			Traces recorded at the numbered points are displayed in panel \formatpanel{B} and \formatpanel{E}.
		\subcaption{B} Example traces recorded at different locations in the parameter space, compare panel \formatpanel{A}.
			The colors of the traces indicate in which compartment the trace was recorded, compare \figEval{}.
			The traces are scaled relative to the height in the first compartment $h_{00}$.
			Due to the faster emulation of the neural dynamics on \gls{bss2}, the time scales are in the microsecond rather than in the millisecond range.
		\subcaption{C} Posterior obtained with the \glsfmtfirst{snpe} algorithm.
			The posterior shows a high density in the parameter region where the target decay constant $\tau^*$ was recorded, \Circled{2}.
			As expected from the grid search result in panel \formatpanel{A}, a correlation between the leak and axial conductance is visible.
			Points where the decay constant is significantly lower/higher than the target observation show a low probability density, \Circled{0} and \Circled{3}.
		\subcaption{D} \num{500} random samples drawn from the approximated posteriors for two different types of observations.
			The green points represent samples drawn from the posterior which is shown in panel \formatpanel{C}.
			The samples show a correlation between both parameters.
			If the absolute heights of the \glsfmttext{psp} which resulted from an input to the first compartment $\myvec{F} = [h_{00}, h_{10}, h_{20}, h_{30}]$ was chosen as observations (blue), the samples scatter around point \Circled{2} where the original target $\myvec{F}^*$ was recorded.
			The histograms at the top and right of the scatter plot show histograms of the parameter distribution in one dimension.
		\subcaption{E} Same traces as in panel \formatpanel{B} but shown on an absolute scale.
			While traces \Circled{1} and \Circled{2} share a similar decay constant $\tau$, compare panel \formatpanel{A} and \formatpanel{B}, their absolute heights differs.
	}
	\label{fig:2d}
\end{figure*}

\subsubsection{Grid Search}\label{sec:grid}
In order to obtain an overview of the model behavior, we performed a grid search over the two-dimensional parameter space.
We created a grid of parameters by choosing equally spaced values of the leak and axial conductance which span the whole parameter range.
The model was then emulated with these parameters on the \gls{bss2} system and the membrane traces in the different compartments were recorded.
In order to easily visualize the results, we selected a one-dimensional observable.
Exponential fits to the maximal height of propagating \glspl{psp} were used in other publications to classify the attenuation of \glspl{psp} in apical dendrites \citep{berger2001high}.
Similarly, we fitted an exponential to the \gls{psp} heights which resulted from an input to the first compartment $\myvec{F} = [h_{00}, h_{10}, h_{20}, h_{30}]$ and analyzed the exponential decay constant $\tau$, \figGridsearch.
The decay constant increased with increasing axial conductance $g_\text{axial}$ and decreasing leak conductance $g_\text{leak}$.
Even though the exponential is just an approximation for the attenuation of transient inputs in multi-compartmental models, a correlation between leak and axial conductance is expected \citep{fatt1951analysis, rall1962electrophysiology}.
This behavior can also be understood with \cref{eq:neuron,eq:mc_current}: a lower leak conductance $g_\text{leak}$ leads to less charge leaking from the membrane and consequently a larger charge transfer to the neighboring compartments, which can be counterbalanced by a lower axial conductance $g_\text{axial}$.

The responses of the membrane potentials to a synaptic input in the first compartment are displayed in \figTracesRelative.
For a low leak and a large axial conductance, \Circled{0}, the attenuation was the weakest and the \gls{psp} was still clearly visible in the last compartment.
Parameters on the same contour line showed, as expected, similar attenuation, \Circled{1} and \Circled{2}, even though the exact shape of the \glspl{psp} differed.
For a large leak and a low axial conductance, \Circled{3}, the \gls{psp} decayed quickly and almost vanished in the third compartment.

\subsubsection{Simulation Based Inference}\label{sec:2d-sbi}
We used the \gls{snpe} algorithm to infer possible parameters $\myvec{\theta} = [g_\text{leak}, g_\text{axial}]$ which reproduce a target observation $\myvec{x}^* = [\tau^*]$.
Furthermore, we investigated how the posterior distribution changed when a more informative observation $\myvec{x}^* = \myvec{F}^* = [h_{00}^*, h_{10}^*, h_{20}^*, h_{30}^*]$ was used, compare \figEval{}.

In the case where a target observation $\myvec{x}^*$ is given by an experiment, the true posterior and the optimal model parameters which replicate the observation are typically unknown.
This makes it hard to assess the quality of the posterior approximated by the \gls{snpe} algorithm.
Therefore, we explicitly chose target parameters $\myvec{\theta}^*$, emulated our model with these parameters on \gls{bss2} and measured an \enquote{artificial} target observation $\myvec{x}^* = \tau^*$.
This allowed us to perform a closure test and check whether the \gls{snpe} algorithm was able to estimate a posterior which agreed with the initial observation.

We picked a target parameter $\myvec{\theta}^*$ at the center of the parameter space and executed the model with this parameter \num{100} times to account for trial-to-trial variations due to temporal noise.
From the full matrix of \gls{psp} heights $\mymat{H}$ we extracted different target observations such as the decay constant $\tau$.
The mean of the observed decay constants was our target observation $\myvec{x}^* = [\overline{\tau}^*] = 1.17 \pm 0.04$; the decay constant is in units of \enquote{compartments}.
In contrast, while running the \gls{snpe} algorithm we executed the model just once for each parameter and did not average over several trials.

We used a uniform distribution over all possible parameters as a prior distribution $p(\myvec{\theta})$ and executed the \gls{snpe} algorithm to obtain an approximation of the posterior distribution $p\left( \myvec{\theta} \mid \myvec{x}^* \right)$.
The uniform distribution covered the whole adjustable range of the leak and axial conductance which ranges from \num{0} to \num{1022}, see \cref{sec:configurability}.

For a number of problems the \gls{snpe} algorithm was reported to be overconfident and ensembles made up of several posteriors were used to retrive a more conservative posterior approximation \citep{hermans2022crisis, deistler2022truncated}.
Since some of your posteriors were also overconfident, see later section, we combined five posterior to a posterior ensemble.

In order to facilitate the comparison of the grid search results and the approximated posterior, we display the difference between the target decay constant $\tau^*$ and the measured decay constant $\tau$ during the grid search in \figGridsearchDiff{}.
As expected form the grid search, a correlation between the leak $g_\text{leak}$ and the axial conductance $g_\text{axial}$ is clearly visible in the approximated posterior, \figPosterior.
The posterior distribution shows high densities for parameters $\myvec{\theta}$ which reproduced observations near the target observation during the grid search.

In order to retrieve a narrow posterior around the original parameters $\myvec{\theta}^*$, a more informative observations was needed.
While the \gls{psp} heights showed a similar decay for different sets of leak and axial conductance, \figTracesRelative, the absolute heights of the \glspl{psp} differed, \figTracesAbsolute.
We therefore used the \glspl{psp} heights which resulted from an input to the first compartment $\myvec{F}$ as a target observation, $\myvec{x}^* = \myvec{F}^*$, to further constrain possible parameters.
The heights in the first compartment $\myvec{F}$ were extracted from the same \num{100} trials as the decay constant $\tau$.
We ran the \gls{snpe} algorithm once again to retrieve another approximation of the posterior.
Samples $\{\myvec{\theta}_i\}$ drawn from this posterior were now scattered around the original parameter $\myvec{\theta}^*$ in the parameter space and the parameters were uncorrelated, \figSamples; the Pearson correlation coefficient decreased from \num{0.92} to \num{0.004}.
The marginal distribution of the leak and axial conductance were bell-shaped and showed a high density near the target parameter $\myvec{\theta}^*$.

\paragraph{Validation}\label{sec:2d_validation}
In order to perform a \gls{ppc}, we drew samples $\{\myvec{\theta}_i\}$ from the posterior distribution, \figPosterior, configured our model with them and compared the observations $\{\myvec{x}_i\}$ with the target observation $\myvec{x}^*$.
We measured a mean decay constant of $\overline{\myvec{\tau}} = 1.18 \pm 0.08$ which agrees with the target $\myvec{\tau}^* = 1.17 \pm 0.04$.
Therefore, we conclude that the approximated posterior is in agreement with the target observation $\myvec{\tau}^*$.
The uncertainty of the posterior predictive increased compared to the target observation since it contains the aleatoric uncertainty, due to the inherent trial-to-trial variations, as well as the epistemic uncertainty which stems from the width of the posterior distribution.

\begin{figure}
	\includegraphics{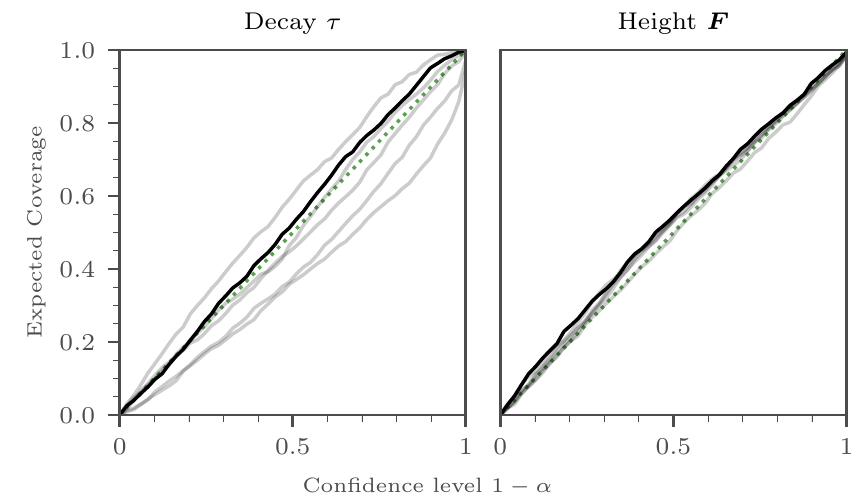}
	\caption{
		Validation of the approximated posteriors found with the  \glsfmtfirst{snpe} for a compartment chain of four compartments and setting parameters globally, compare \cref{fig:2d}.
		The gray lines mark the expected coverage \citep{hermans2022crisis, deistler2022truncated} of posterior approximations found with the \glsfmttext{snpe} algorithm, the black line marks the expected overage of the posterior ensemble which is made up of those posteriors.
		Left: Using the decay constant $\tau$ as an observable.
		Several posteriors have an expected coverage below the diagonal which indicates an overconfident probability distribution.
		When combining several posteriors to an ensemble, the expected coverage follows the diagonal which is a sign of a well calibrated posterior.
		Right: \Acrlong{psp} heights resulting from an input to the first compartment as an observation.
		All posteriors and their ensemble appear well calibrated as they closely follow the diagonal.
	}
	\label{fig:2d_validation}
\end{figure}

To test the calibration of the approximated posterior, we calculated the expected coverage, compare \cref{sec:validation}.
When we used the decay constant $\tau$ as a target, three out of five posteriors were overconfident, \cref{fig:2d_validation}.
We followed the methods presented in \citet{hermans2022crisis, deistler2022truncated} and combined five posteriors to form an ensemble.
The expected coverage of this ensemble closely follows the diagonal and indicates a well calibrated posterior.

In case of the heights $\myvec{F}$ as a target, the single posteriors were already well calibrated.
And consequently, the ensemble of five posteriors was also well calibrated.

In the appendix, we compare the results from the emulation on \gls{bss2} with computer simulations performed in the simulation library Arbor \citep{abi2019arbor}, \cref{fig:arbor}.

\subsection{Multidimensional Parameter Space}\label{sec:multi}

\begin{figure*}
	\includegraphics{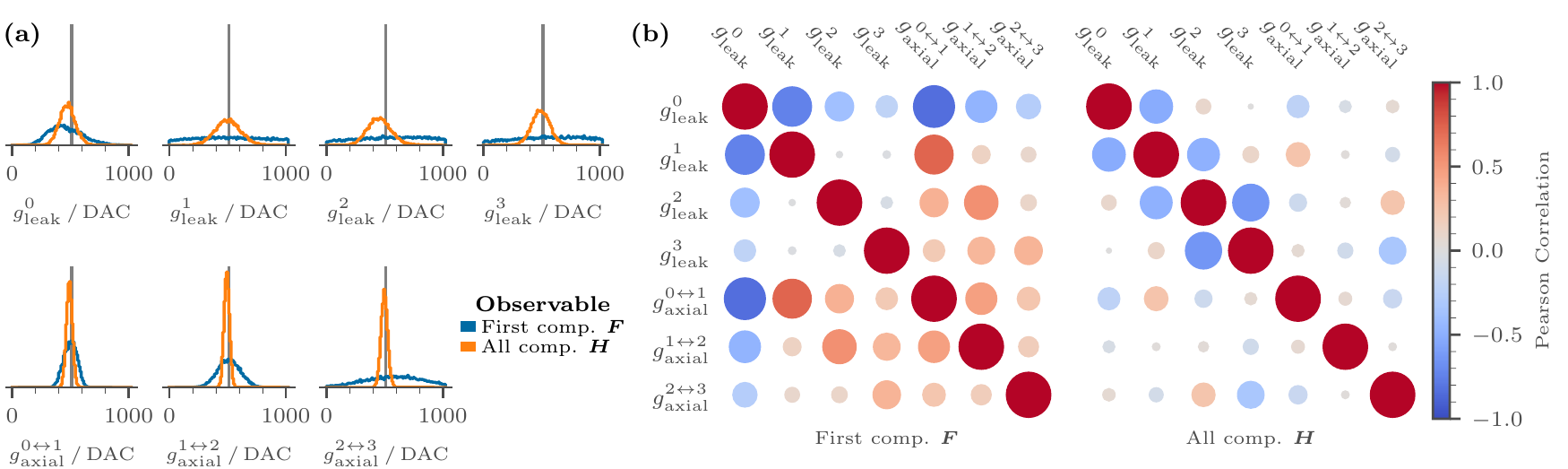}
	\caption{
		Results of the \glsfmtlong{snpe} algorithm for a compartment chain of \num{4} compartments and setting parameters individually for each compartment and connection between them.
		Emulations were performed on the neuromorphic \glsfmtlong{bss2} system.
		\subcaption{A} Histograms of \num{10000} parameters drawn from the approximated posterior.
			For the heights $\myvec{F}$ of the \glsfmtfirstpl{psp} which resulted from an input to the first compartment as a target observation (blue), the distribution of the leak conductance in the first compartments is bell-shaped and peaks near the target parameter (dotted line).
			The leak conductance is roughly uniformly distributed in later compartments.
			The distributions of the axial conductance are bell-shaped and broaden for later compartments.
			Choosing all heights $\mymat{H}$ as a target (orange) leads to narrower distributions.
			All histograms are now bell-shaped with a peak near the target (dotted line).
		\subcaption{B} Pearson correlations between different parameters.
		The color denotes the value of the correlation while the radius of the circle encodes the absolute value of the correlation.
		Left: \glsfmttext{psp} heights resulting from an input to the first compartment $\myvec{F}$ as a target.
		The strongest correlations can be observed for the leak conductance in the first compartment $g_\text{leak}^0$ and the axial conductance between the first and second compartment $g_\text{axial}^{1 \leftrightarrow 2}$; for parameters later in the chain the correlations shows lower values.
		Right: All \glsfmttext{psp} heights $\mymat{H}$ as a target.
		Overall the correlations decrease for this more informative target.
		Only between neighboring leak conductances a high negative correlation can be observed.
	}
	\label{fig:md}
\end{figure*}

In order to increase the problem complexity, we set the leak and axial conductance for each compartment and connection individually.
For four compartments this resulted in a total of seven parameters; four leak condutances $g_\text{leak}^i (i=0, 1, 2, 3)$ and three axial conductances $g_\text{axial}^{i \leftrightarrow i + 1} (i=0, 1, 2)$.

As in the previous section we used a uniform prior and the \gls{psp} heights caused by an input to the first compartment as a target ($\myvec{x}^*=\myvec{F}^*$).
We then executed the \gls{snpe} algorithm, combined five approximated posteriors to an ensemble and drew samples from this posterior $p(\myvec{\theta} \mid \myvec{x}^*)$.

The marginal distribution of the sampled leak conductance in the first compartment $g_\text{leak}^0$ was bell-shaped and peaked near the target parameter, \figMarginals.
The almost uniform distributions of the leak conductances in the other compartments indicated that they were not relevant for the chosen observation.
In contrast, the marginal distribution of all axial conductances were bell-shaped with a high density around the original parameters.
The distributions of the axial conductance became broader for conductances later in the chain, suggesting that the influence of these conductances on the observable was weaker.

Similar to the two-dimensional case, we considered a higher-dimensional observation as a target to retrieve narrower posterior distributions, i.e., we chose all \gls{psp} heights as a target ($\myvec{x}^* = \mymat{H}^*$), \figMarginals.
Now the one-dimensional marginals of all parameters were bell-shaped.
The marginals of the axial conductance showed a narrower distribution than these of the leak conductance, indicating that the given observation was more sensitive to the axial conductance.

\paragraph{Correlation}
In \figCorrelations{} we display the correlation between posterior samples, the one- and two-dimensional marginals of posterior samples can be found in \cref{fig:pairplot_hd_amp_first,fig:pairplot_hd_amp_all}.
When we considered the \gls{psp} heights which resulted from an input to the first compartment $\myvec{F}$ as an observable we saw strong negative correlations between the leak conductance in the first compartment $g_\text{leak}^0$ and the leak conductance in the neighboring compartment $g_\text{leak}^1$ as well as the axial conductance between both compartments $g_\text{axial}^{1 \leftrightarrow 2}$.
This can be explained with \cref{eq:neuron,eq:mc_current} and considering the \gls{psp} height in the first compartment: when the leak conductance $g_\text{leak}^0$ in the first compartment increases, a higher current leaks from the membrane which would result in a smaller \gls{psp} height; to counter this effect the charge which flows to the neighboring compartment has to be minimized by reducing the axial conductance $g_\text{axial}^{1 \leftrightarrow 2}$ between the compartments or the leak conductance $g_\text{leak}^1$ of the neighboring compartment.

The leak conductance $g_\text{leak}^0$ was also negatively correlated to the other leak and axial conductances, compare first column in \figCorrelations{}.
The magnitude of the correlation decreased for parameters further away from the first compartment.
Apart from the correlation with the leak conductance $g_\text{leak}^0$ of the first compartment, the correlation between the other leak conductances was low.

Interestingly, the correlations between the axial conductances $g_\text{axial}^{i \leftrightarrow i+1}$ and the other leak conductance $g_\text{leak}^i, i>0$ was positive.
As mentioned above a higher leak conductance leads to a larger leak current which results in a smaller \gls{psp} height.
Since we only considered an input to the first compartment, this increased leak conductance $g_\text{leak}^i$ could be counteracted by increasing the charge which is injected from the previous compartments and therefore increasing the conductance $g_\text{axial}^{j-1 \leftrightarrow j}; j\le i, i>0$ to compartments earlier in the chain.
The correlation of the leak conductance $g_\text{leak}^i$ to later axial conductances $g_\text{axial}^{j \leftrightarrow j+1}; j\ge i, i>0$ were still positive but significantly lower.

As expected, all axial conductances were correlated positively.
This can be explained when considering one compartment $i, i>0$.
An increase in the axial conductance $g_\text{axial}^{i-1 \leftrightarrow i}$ leads to a stronger current on compartment $i$; to prevent an accumulation of charge and therefore a larger \gls{psp} height, the conductance to the next compartment $g_\text{axial}^{i \leftrightarrow i+1}$ has to increase as well.

When taking all heights $\myvec{H}$ as a target observation, the correlation between the different parameters decreased.
Only between neighboring leak conductances $g_\text{leak}^i$ and $g_\text{leak}^{i+1}$ rather high negative correlations could be observed.
To understand this correlation, we can consider two cases.
We look at one compartment $i$, increase its leak conductance $g_\text{leak}^i$ and consider once an input to the same compartment $i$ and once to a neighboring compartment $i\pm1$.
First, input in the same compartment: as before an increased leak conductance $g_\text{leak}^i$ results in a lower \gls{psp} height which can be compensated by a smaller leak conductance $g_\text{leak}^{i\pm1}$ in the neighboring compartments.
Similarly, in the second case when the input is injected in a neighboring compartment, an increased leak conductance $g_\text{leak}^i$ would once again lead to an increased leakage and a decreased \gls{psp} height.
Consequently, the leak conductance $g_\text{leak}^{i\pm1} $ in the neighboring compartment should be reduced such that more charge can flow on compartment $i$.

\paragraph{Validation}
We once again used \glspl{ppc} to check if samples drawn from the approximated posterior $\{\myvec{\theta}_i\}$ reproduce the target observation.
The mean difference between observations $\{\mymat{H}_i\}$ obtained with these parameters and the target observation $\mymat{H}^*$ are displayed in \figPPC.
$\mymat{H}$ describes all observed \gls{psp} heights and the target observation $\myvec{F}^*$ was extracted from $\mymat{H}^*$, see \figEval{}.

The mean of the \gls{psp} heights for an input to the first compartment (first column) was near the initial target values; the standard deviation was in the range of \numrange{1}{2}~$\sigma^*$ where $\sigma^*$ is the standard deviation of the measurements which were used to extract the target observation $\mymat{H}^*$.
For responses in the first compartment (first row) a similar standard deviation could be observed, but the mean observation showed a slightly higher deviation from the target observation.
For the other \gls{psp} heights the mean was still in the one-sigma range of the initial target observation, but the standard deviation of the observations was significantly higher.
The small deviation of the mean observations can be explained by our target parameter which is located at the center of the parameter space; a prior predictive check also yielded mean observations near the target observations, compare \cref{fig:prior_predictive}.
The higher standard deviations are expected since these \gls{psp} heights have not been part of the observation and can be attributed to the broad posterior distribution of the leak and axial conductance in later compartments, compare \figMarginals{}.

The sharpening of the posterior distribution was also visible in the results of the \gls{ppc}, \figPPC.
Here the standard deviation of the observations decreased to the range of \numrange{1}{2}~$\sigma^*$ for all \gls{psp} heights.

As for the two-dimensional case, we calculated the expected coverage for the approximated posteriors and their ensembles, \figCoverageMd{}.
With the heights which resulted from an input to the first compartment $\myvec{F}$ as an observable, all five posteriors were overconfident and also an ensemble made up of these five posteriors was still overconfident.
When using all heights $\mymat{H}$ as an observation, the expected coverage of the individual posteriors were similar to the case before.
However, the expected coverage of the ensemble was near the diagonal; this suggests that the posterior was well calibrated.

\begin{figure*}
	\includegraphics{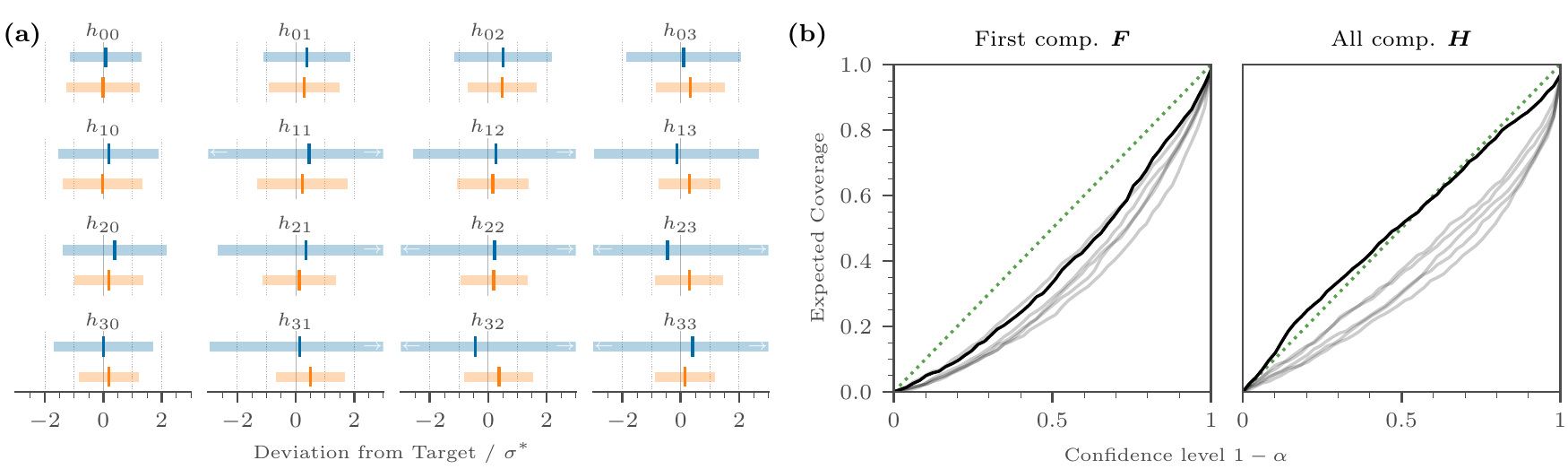}
	\caption{
		Validation of the approximated posteriors found with the  \glsfmtfirst{snpe} for a compartment chain of four compartments and setting parameters individually for each compartment and connection between them, compare \cref{fig:md}.
		Emulations were performed on the neuromorphic \glsfmtlong{bss2} system.
		\subcaption{A} \Acrlong{ppc}.
			The passive chain was configured with \num{1000} of the parameters $\{\myvec{\theta}_i\}$ drawn in \figMarginals{} and the \glsfmtfirst{psp} heights in all compartments $\{\mymat{H}_i\}$ were measured on the \glsfmtlong{bss2} system.
			These \glsfmttext{psp} heights were compared to the observation $\mymat{H}^*$ which represents the measurement with the target parameters $\myvec{\theta}^*$.
			The vertical lines show the mean deviation of the observations $\{\mymat{H}_i\}$ from this target $\mymat{H}^*$ while the horizontal bars illustrate the standard deviation of this deviation.
			As mentioned in the introduction, analog hardware is subject to temporal noise.
			Therefore, the hardware was configured to the target parameters $\myvec{\theta}^*$ \num{100} times and the mean \glsfmttext{psp} heights were chosen as a target $\mymat{H}^*$; the deviations in this panel are scaled by the standard deviation $\sigma^*$ of these \num{100} measurements (each height deviation $h_{ij}$ is divided by the standard deviation of the height $\sigma^*_{ij}$).
			For all \glsfmttext{psp} heights the mean observation is within \numrange{1}{2} standard deviations of the initial target.
			When a more informative observation $\mymat{H}$ is chosen, the standard deviations decreases.
			A prior-predictive check can be found in the appendix, \cref{fig:prior_predictive}.
		\subcaption{B} Coverage tests. The gray lines mark the expected coverage \citep{hermans2022crisis, deistler2022truncated} of posterior approximations found with the \glsfmttext{snpe} algorithm, the black line marks the expected overage of the posterior ensemble which is made up of these posteriors.
		Left: \glsfmttext{psp} heights resulting from an input to the first compartment $\myvec{F}$ as an observation.
		The expected coverage is for all confidence levels below the diagonal which suggests that the posteriors are overconfident.
		Even an ensemble made up of five posteriors is not well calibrated.
		Right: All \glsfmttext{psp} heights $\mymat{H}$ as a target.
		The individual posteriors are overconfident but the ensemble of them is well calibrated.
	}
	\label{fig:md_validation}
\end{figure*}
 	\section{Discussion}
\glsresetall

We have shown that the \gls{snpe} algorithm can be used to parameterize the analog neuromorphic \gls{bss2} system.
To be able to investigate the posteriors approximated by the \gls{snpe} algorithm, we selected a multi-compartmental model which takes the form of a chain of passive compartments.
We chose the leak conductance as well as the axial conductance between compartments as parameters and observed how \glspl{psp} propagated along the chain.
This model allowed us to easily change the dimensionality of the parameter space as well as the choice of observable and evaluate how this influences the approximated posteriors.

In all our experiments, we picked a set of target parameters, extracted an observation with these parameters and then used the \gls{snpe} algorithm to approximate the posterior distribution of the parameters which reproduce this given observation.

As a first step, we considered a two-dimensional parameter space where we set all leak conductances and axial conductances to the same value.
The low dimensionality of the parameter space allowed us to perform a grid search in a reasonable amount of time.
The posterior approximated by the \gls{snpe} algorithm agreed with the results from this grid search.
In both cases we found a correlation between the leak and axial conductance when looking at the attenuation of \glspl{psp}; this argees with theoretical expectations \citep{fatt1951analysis, rall1962electrophysiology}.
To be able to find such correlation is one of the advantages of a posterior approximation over traditional parameter search algorithms which usually only yield a set of parameters which reproduce the given observation but do not illustrate the relation between different parameters.

When we chose a more informative observation, specifically the height of the \glspl{psp} which result from an input to the first compartment, the posterior distribution of the parameters narrowed and the correlation between leak and axial conductance vanished.
We further showed that the algorithm is capable of finding appropriate posterior approximations for several, random values of the target parameters.
The approximations were even in agreement with the target parameters if they lie at the edges of the parameter space.
This indicates that the algorithm is able to deal with the hard parameter limits which are dictated by the neuromorphic hardware.

Furthermore, we performed coverage tests to assess the calibration of the posterior approximations.
The posteriors produced with the less informative observation were overconfident, requiring an ensemble of five posteriors to retrieve a well calibrated posterior.
In contrast, all posteriors approximated for the more informative observation were well calibrated.

Next, we increased the dimensionality of the parameter space by adjusting each leak and axial conductance individually; resulting in a seven-dimensional parameter space.
We showed that the marginal distributions of samples drawn from the posterior approximations have a high density around the target parameters.
In addition, we analyzed the correlation between the different parameters and showed that they agree with the model equations.

Furthermore, we conducted \glspl{ppc} to verify that the parameters drawn from the approximated posterior yield emulated results which align with the target observation.
Similar to the two-dimensional case, increasing the dimensionality of the observable resulted in a narrower posterior distribution.
When using the height of the \glspl{psp} which resulted from an input to the first compartment as an observable, we did not find well calibrated posteriors even when combining multiple posteriors into an ensemble.
After increasing the dimensionality of the observable, the individual posteriors remained overconfident but the ensemble made up of five of them was well calibrated.
 	\section{Conclusion}\label{sec:conclusion}
The \gls{snpe} algorithm has previously only been utilized to identify suitable parameters for numerical simulations \citep{lueckmann2017flexible, greenberg2019automatic, goncalves2020training, deistler2022truncated}.
In the current work we show that the algorithm can also be employed to parameterize a physical system, namely the \gls{bss2} neuromorphic system.

In contrast to other search algorithms such as random search, genetic algorithms or gradient-based algorithms, the \gls{snpe} algorithm provides an approximation of the full posterior and therefore allows to identify correlations between parameters and to quantify the confidence of the parameter estimation.
Additionally, the \gls{snpe} algorithm is agnostic to the internal dynamics of the experiment and does not require the calculation of gradients.
Compared to traditional \gls{sbi} methods the \gls{snpe} algorithm offers a higher simulation efficiency \citep{papamakarios2016fast, cranmer2020frontier}.
As a result, \gls{snpe} is a viable alternative to traditional optimization methods.

When one simply optimizes for a single objective and is not concerned with the correlations between parameters, gradient based methods can offer a more directed optimization approach and are potentially faster in recovering suitable parameters; they have successfully been used to find parameters for \gls{bss2} \citep{cramer2022surrogate, pehle2023event, arnold2023spiking}.
However, having access to an approximated posterior distribution and the correlations between different parameters can give valuable insight in the dynamics of the underlying model as shown in the current study.

To evaluate the quality of the approximated posteriors, we generated the target observation from our model.
As a result, we knew the true parameters of the target observation and were certain that our model can reproduce the given observation.
In subsequent studies, we will use the \gls{snpe} algorithm to replicate observations which are generated by another model such as numerical simulations or by physiological experiments.

Furthermore, we only considered passive neuron properties in our current experiments.
As seen in the grid search results, this lead to a rather smooth parameter space, where the observations change gradually with the model parameters.
More complex neuron models of interest include non-linear behavior such as somatic or dendritic spikes and will potentially have high-dimensional parameter spaces.
\citet{goncalves2020training, deistler2022truncated} have previously shown that the \gls{snpe} algorithm and derivatives of it can deal with such high-dimensional parameter spaces and non-linear behavior and it will be interesting if this success can be transferred to emulations on neuromorphic hardware.

In summary, we demonstrated that the \gls{snpe} algorithm is able to find posterior approximations for parameters of the analog neuromorphic \gls{bss2} system.
 
	\section*{Contributions}
	J.K., J.S. and S.S. designed research; J.K. and R.S. performed research; J.K., J.S, R.S. and S.S. analyzed data; J.K. and S.S. wrote the paper; all authors edited the paper; E.M., J.K., R.S. and S.S contributed software; and J.S. designed the BrainScaleS-2 neuromorphic system.
 
	\section*{Acknowledgments}
	We thank the lead of HBP's FIPPA project Christian Tetzlaff for scientific input; A. Baumbach, S. Billaudelle, A. Gr\"ubl, J. Ilmberger, C. Mauch, C. Pehle, Y. Stradmann, P. Spilger and, J. Weis for their contributions to \gls{bss2}; as well as all present and former members of the Electronic Vision(s) research group.
We thank the anonymous reviewers for their thorough evaluation and valuable suggestions.

This work has received funding from the EU ([FP7/2007–2013], [H2020/2014–2020]) under grant agreements 604102 (HBP), 720270 (HBP SGA1), 785907 (HBP SGA2) and 945539 (HBP SGA3); the Deutsche Forschungsgemeinschaft (DFG, German Research Foundation) under Germany’s Excellence Strategy EXC 2181/1-390900948 (the Heidelberg STRUCTURES Excellence Cluster) as well as from the Manfred St\"ark Foundation.
 
	\section*{Data Availability}
	The data that support the findings of this study are openly available \citep{kaiser2023simulation_data}.
The experiment code is available on \url{https://github.com/electronicvisions/model-paper-mc-sbi}.

	\printbibliography

	\appendix
	\section{Appendix}
\renewcommand{\thefigure}{A\arabic{figure}}
\setcounter{figure}{0}
\glsresetall

\begin{figure}
	\includegraphics{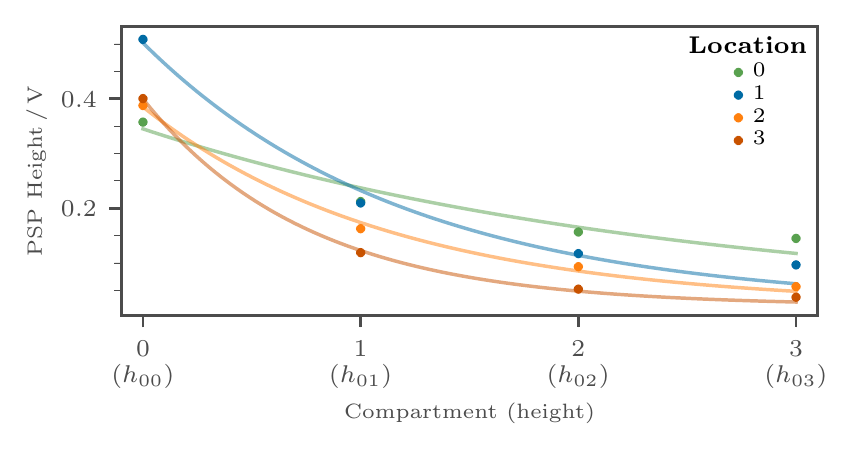};
	\caption{Exponential fit to the traces displayed in \figTracesRelative{} and \figTracesAbsolute{}.
			 The heights of the \glsfmtfullpl{psp} are extracted from the recorded membrane traces, compare \figEval{}, and exponentials (solid lines) are fitted to the measurement points.
			 The numbering is the same as in \cref{fig:2d}.
			 The x-axis label mark the compartment in which the height of the \glsfmttext{psp} was measured and in brackets the variable name as defined in \figEval{}.
			 }
	\label{fig:exponential}
\end{figure}

\begin{figure*}
	\includegraphics{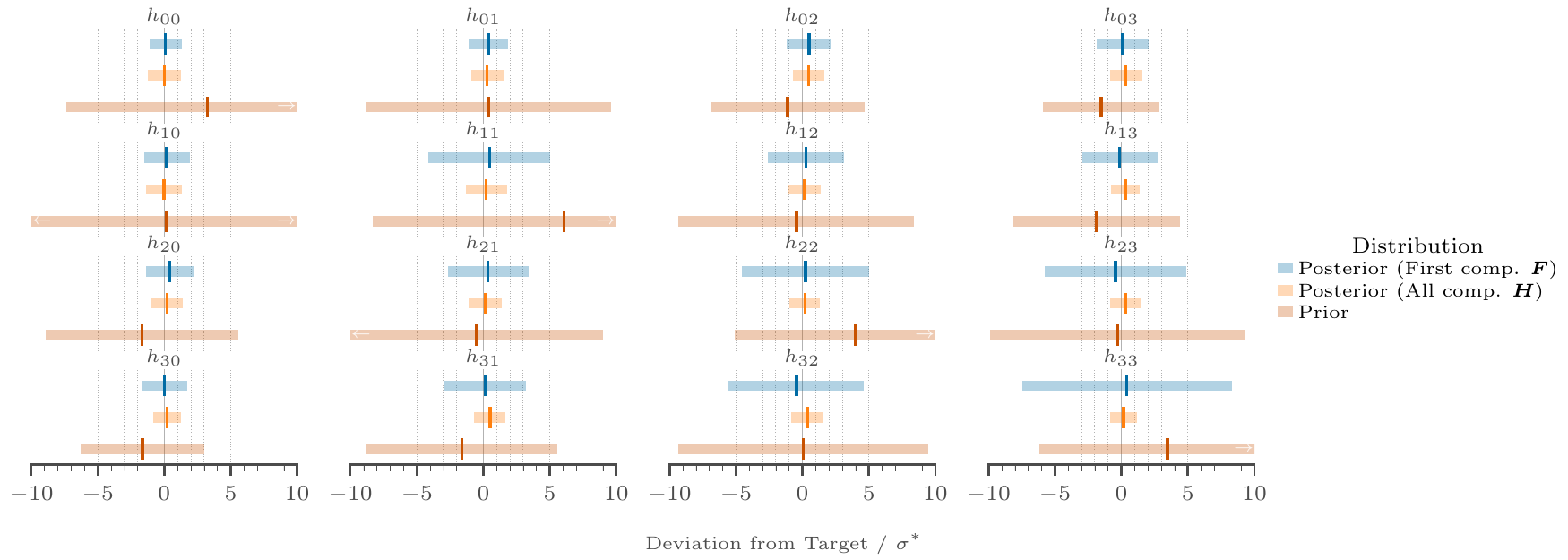};
	\caption{
		Comparison between a prior-predictive check and the \glsfmtfirstpl{ppc} in \figPPC{}.
		The data for the \glspl{ppc} are copied from \figPPC{}, the \gls{ppc} was performed for two different observations: \glsfmtfirst{psp} heights which resulted from an input to the first compartment $\myvec{F}$ and all \glsfmttext{psp} heights $\mymat{H}$.
		The prior-predictive check was performed similar to the \glspl{ppc} but the samples were drawn from the prior distribution $p(\myvec{\theta})$.
		The vertical lines show the mean deviation of the observations $\{\mymat{H}_i\}$ from this target $\mymat{H}^*$ while the horizontal bars illustrate the standard deviation of this deviation.
		The hardware was configured to the target parameters $\myvec{\theta}^*$ \num{100} times and the mean \glsfmttext{psp} heights were chosen as a target $\mymat{H}^*$; the deviations in this figure are scaled by the standard deviation $\sigma^*$ of these \num{100} measurements (each height deviation $h_{ij}$ is divided by the standard deviation of the height $\sigma^*_{ij}$).
		While the mean observation is not so far off from the target observation in most cases, the standard deviation is significantly higher than for the \glspl{ppc}.
		 }
	\label{fig:prior_predictive}
\end{figure*}

\begin{figure*}
	\includegraphics{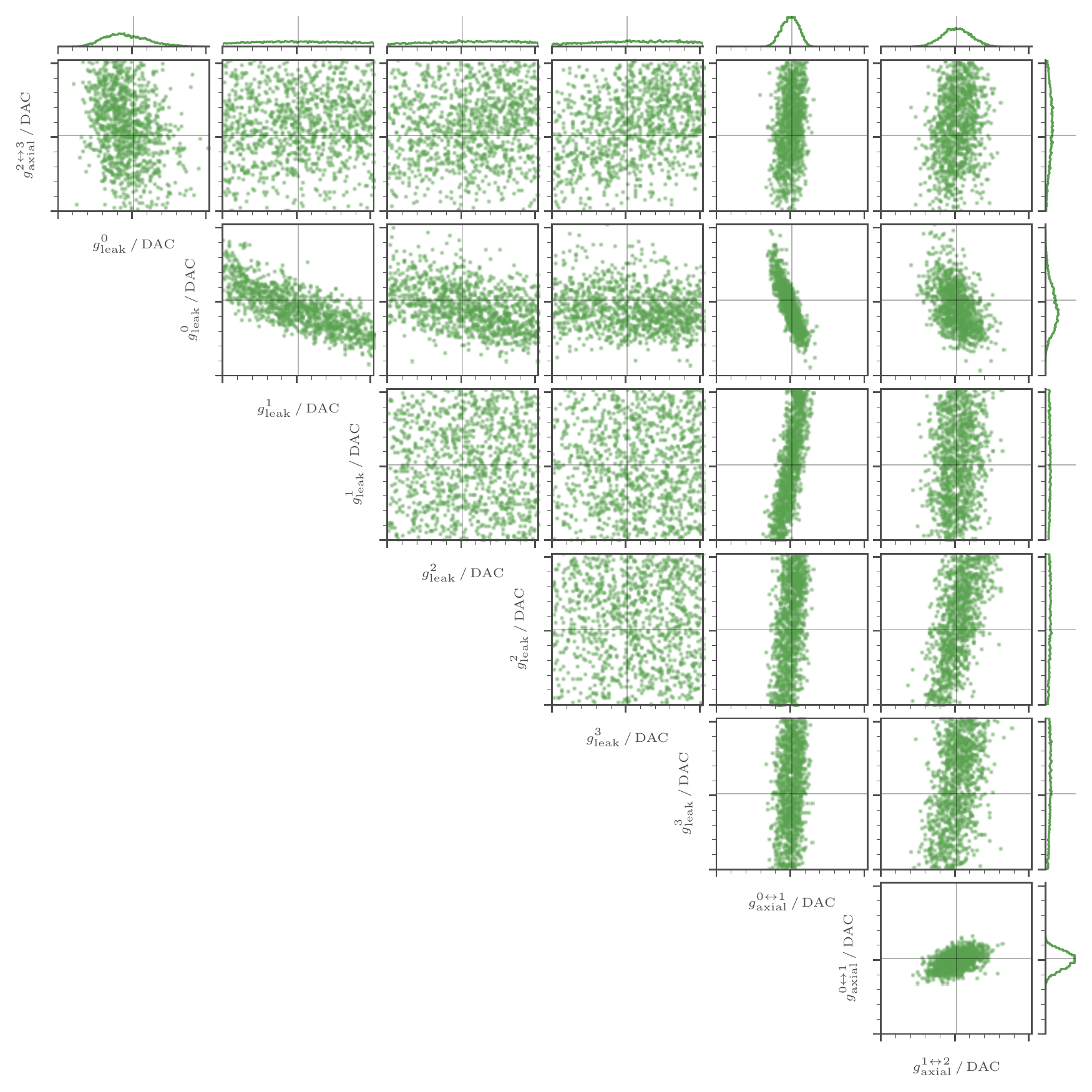};
	\caption{
		One- and two-dimensional marginal distributions of \num{1000} samples which were drawn from the posterior ensembles displayed in \cref{fig:md,fig:md_validation}, the target observations were the \glsfmttext{psp} heights $\myvec{F}$ which resulted from an input to the first compartment.
			 }
	\label{fig:pairplot_hd_amp_first}
\end{figure*}

\begin{figure*}
	\includegraphics{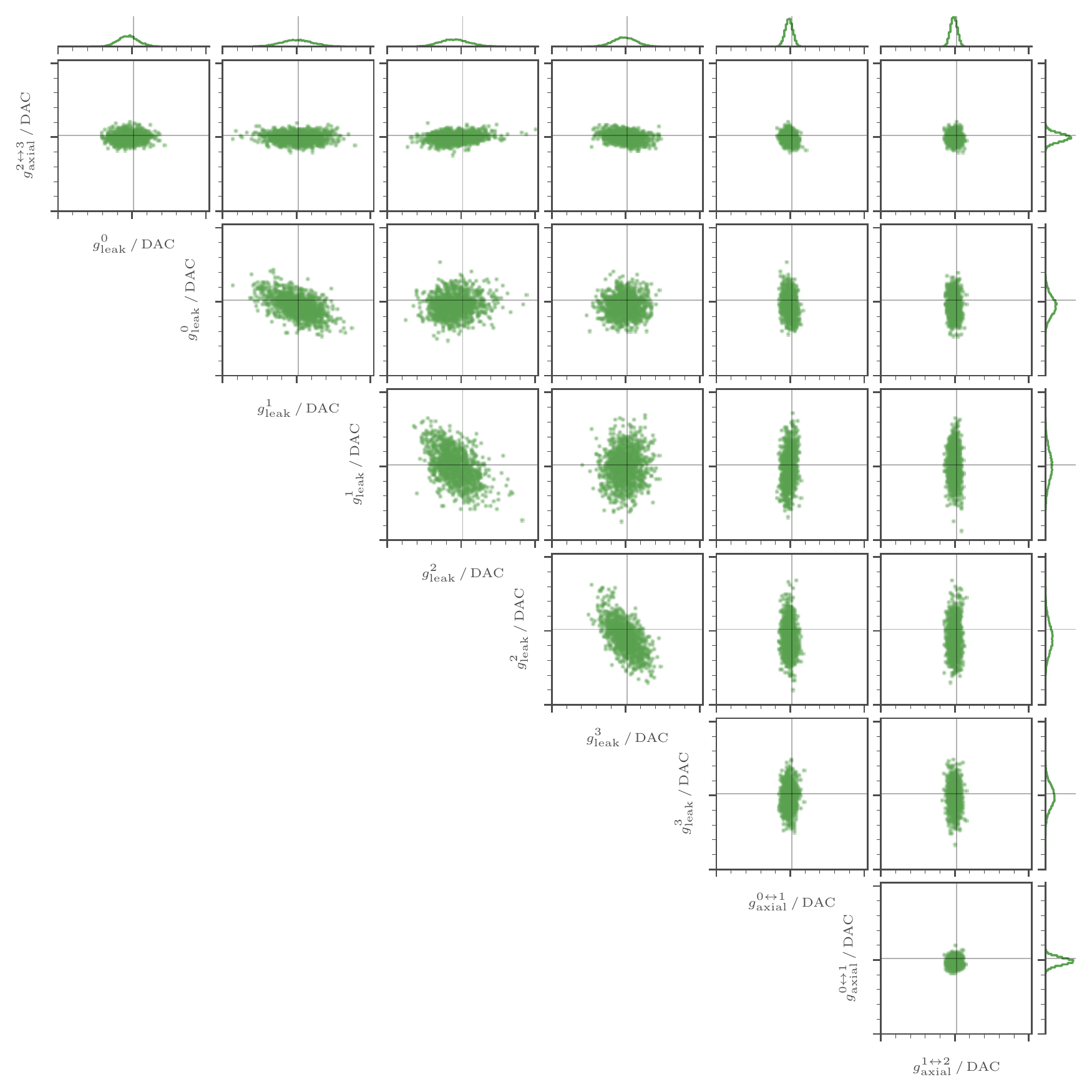};
	\caption{
		One- and two-dimensional marginal distributions of \num{1000} samples which were drawn from the posterior ensembles displayed in \cref{fig:md,fig:md_validation}, the target observation were all \glsfmttext{psp} heights $\mymat{H}$.
			 }
	\label{fig:pairplot_hd_amp_all}
\end{figure*}

In the following appendix, we state which neuron parameters were used during the experiment, how the hyperparameters of the \gls{snpe} algorithm influenced the approximated posterior and compare our results to simulations in Arbor \citep{abi2019arbor}.
Furthermore, we attach figures which extend the results presented in the paper; \cref{fig:exponential,fig:prior_predictive,fig:pairplot_hd_amp_first,fig:pairplot_hd_amp_all}.

\subsection{Neuron Parameters}
In order to ensure a similar behavior of the different compartments, the leak potential and the synaptic properties were calibrated.
The synaptic time constant was calibrated to a value of \SI{10}{\us}.
As can be extracted from \figGridsearch{}, the decay constant varied in our experiments between \numrange{0.16}{4.08} compartments.
When varying the leak conductance $g_\text{leak}$ over the full range specified in \cref{fig:2d}, the membrane time constant $\tau_\text{m} = \frac{C_\text{m}}{g_\text{leak}}$ varies in the range of \SIrange{12}{30}{\us}.

\subsection{Sequential Neural Posterior Estimation Algorithm}\label{sec:meth:snpe}
We adjusted the number of simulations as well as the properties of the \gls{nde} and used \glspl{ppc} to check how these hyperparameters influence the approximated posterior.
For each set of hyperparameters we executed the \gls{snpe} algorithm ten times with different seeds.
The seeds influence the initial weights as well as the parameters $\myvec{\theta}$ which are drawn from the prior in the first round.
Different sets of hyperparameters shared the same seeds.

\begin{figure}
	\includegraphics{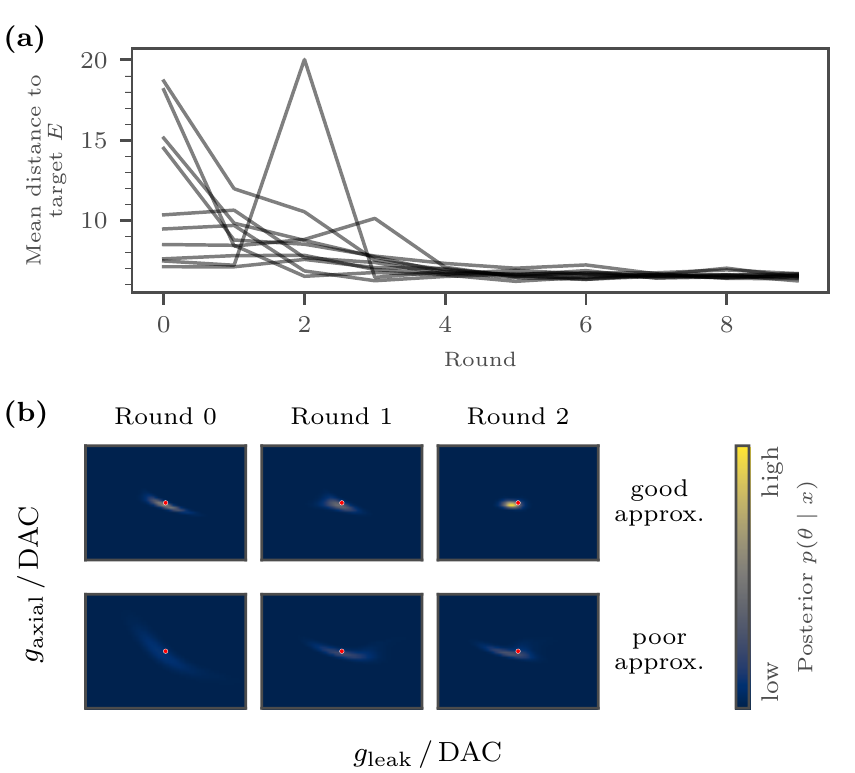}
	\caption{
		Evolution of the approximated posterior over several rounds of the \glsfmtfirst{snpe} algorithm.
		Results are shown for emulations executed on the \glsfmtlong{bss2} system.
		\subcaption{A} \Glsfmtfirst{ppc} for a emulation budget of \num{10} rounds with \num{50} emulations in each round (the \glsfmttext{ppc} was executed with \num{1000} parameters sampled from the posteriors).
			The \glsfmttext{snpe} algorithm was executed \num{10} times with different seeds.
			For some executions of the \glsfmttext{snpe} algorithm, the approximated posterior in the first round poorly replicates observations which are similar to $\myvec{x}^*$; this is evident in a high mean distance $E$.
			In all displayed cases the \glsfmttext{snpe} algorithm is able to recover a meaningful posterior.
		\subcaption{B} Examples for one case where the \glsfmttext{snpe} algorithm is able to approximate a meaningful posterior and one case in which the algorithm fails to find a good approximation in the frist three rounds.
			In both cases, the approximation in the first round does not agree with the true posterior.
			In the top row, the algorithm is able to quickly recover from the poor approximation while in the bottom row more rounds are needed to obtain a meaningful approximation.
			The parameter ranges are the same as in \figPosterior{}.
	}
	\label{fig:simulations}
\end{figure}

\subsubsection{Number of Simulations and Rounds}
For the two-dimensional parameter space and the decay constant $\tau$ as an observable, three times \num{50} emulations were sufficient to recover a posterior which is in agreement with the target observation.
Retrieving the observation of a single emulation (including hardware configuration, experiment execution, data retrieval and evaluation) took about took about \SI{420}{\ms}.

When the observable is changed to the height of the \glspl{psp} which result from an input to the first compartment $\myvec{F}$, the \gls{snpe} algorithm failed to find a suitable approximation if the number of emulations was too low.
This was due to a poor approximation in the first round from which the algorithm needed some time to recover or may not recover in the given emulation budget, \cref{fig:simulations}.
We observed that a higher number of emulations in the first round reduced the number of cases where the posterior was approximated poorly.
Therefore, we chose \num{500} emulations in the first round followed by ten rounds of \num{50} emulations for a two-dimensional parameter space with $\myvec{F}$ as an observable.
We used two times \num{1000} emulations for the multidimensional parameter space, \cref{sec:multi}.

\begin{figure}
	\includegraphics{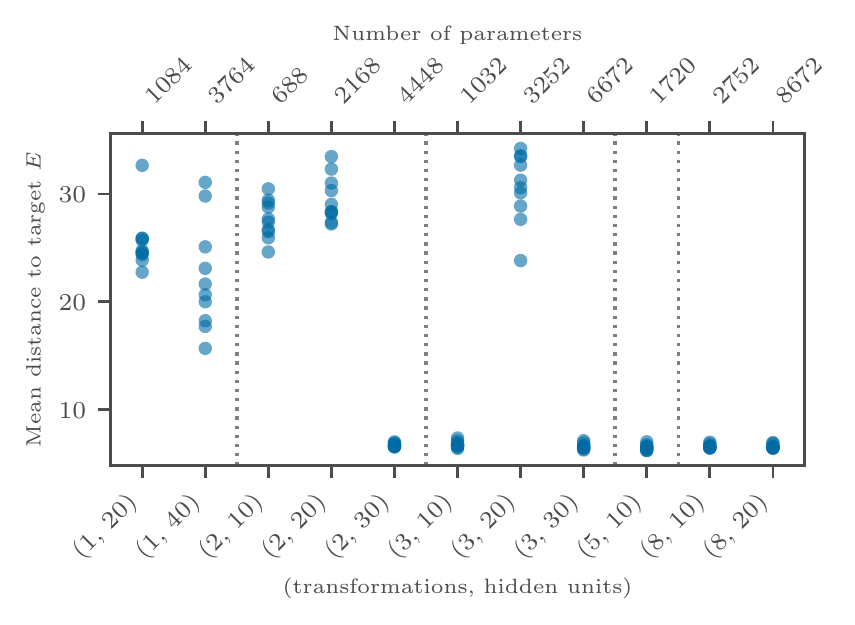}
	\caption{Influence of the parameterization of the \glsfmtfirst{nde} on the approximation of the posterior.
		We used \glsfmtfirstpl{maf} as \glsfmtshortpl{nde}.
		\glsfmtshortpl{maf} transform normal distributions in other distributions \citep{papamakarios2017masked}.
		We used transformations which are made up of two blocks and change the number of hidden units which are used in each block \citep{goncalves2020training}.
		Furthermore, we changed the number of transformations which are chained together.
		As in \figPosteriorEvo{} we performed a \glsfmtlong{ppc} and used the mean distance between these samples and the target as a measure to decide if the approximation agreed with the target observation $\myvec{x}^*$.
		Again, we used the \glsfmtlong{psp} heights resulting from an input to the first compartment as an observable and repeated the \glsfmtlong{snpe} algorithm with \num{10} different seeds for each set of hyperparameters.
		At least two transformation were needed to recover a meaningful posterior.
		The number of experiments in which a meaningful posterior could be recovered seemed to increase with the number of transformations.
		The total number of trainable parameters was not an indicator how well the \glsfmttext{nde} was able to approximate the true posterior.
	}
	\label{fig:transforms}
\end{figure}

\subsubsection{Neural Density Estimator}
Based on the results in \citet{lueckmann2021benchmarking} we use \glspl{maf} as \glspl{nde} \citep{papamakarios2017masked}.
\Glspl{maf} transform normal distributions in other probability distributions.
We used the values provided by the \texttt{sbi} package \citep{tejero2020sbi} as defaults; similar values have also been used in previous publications \citep{lueckmann2021benchmarking, goncalves2020training}.
Here the \gls{maf} is made up of five transformations which are chained together.
Each of these transformation consists of two blocks with \num{50} hidden units per block.
For more information see \citet{papamakarios2017masked,papamakarios2021normalizing}.

In case of a two-dimensional parameter space and the decay constant $\tau$ as a target, \cref{sec:2d}, a single transformation with two blocks of ten hidden units each was sufficient.
If we selected the heights which result from an input to the first compartment $\myvec{F}$ as a target, a single transformation was not sufficient to recover a meaningful posterior, \cref{fig:transforms}.
Starting from two transformations and \num{30} hidden units, the best value of the \gls{ppc} were obtained.
The only exception is the network with three transformations and \num{20} hidden units for which the algorithm could not recover from a poor approximation in the first round.

A \gls{maf} with one transformation and \num{50} hidden units is made up of \num{3764} trainable parameters and fails to approximate the true posterior.
On the other hand a \gls{maf} with five transformations and \num{10} hidden units in each block offers just \num{1720} trainable parameters but is able to find approximations which agree with the target observation.
We conclude, that a high number of transformations was more important for a good posterior approximation than a high number of trainable parameters.
For the results reported in \figSamples{} we used the \gls{nde} with five transformations, two blocks and ten hidden units.

\subsection{Choice of the Target Parameters}

\begin{figure}
	\includegraphics{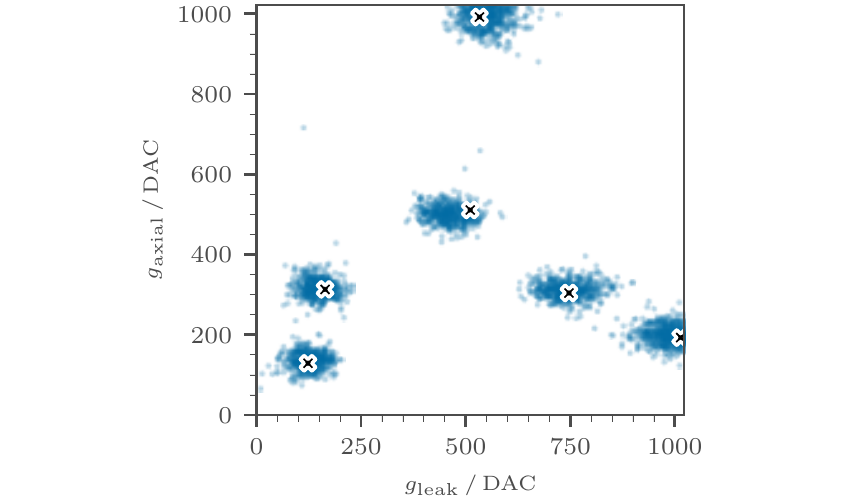}
	\caption{Posterior samples $\{\myvec{\theta}_j\}_i \sim p(\myvec{\theta} \mid \myvec{x}^*_i)$ for different observations $\myvec{x}^*_i$.
			We drew five random parameters ${\myvec{\theta}_i}$ from a uniform prior and one parameter at the center of the parameter space (marked by black crosses).
			The target observations $\{\myvec{x}^*_i\}$ were obtained by emulating the model \num{100} times for each parameter on \glsfmtlong{bss2} and taking the mean height of the \glsfmtlong{psp} obtained from an input to the first compartment, compare \figSamples.
			As a posterior approximation we used the first round posterior obtained while executing the \glsfmtlong{snpe} algorithm in \cref{sec:2d-sbi}.
			The samples drawn from the approximated posterior (small dots) are in the vicinity of the parameters which were used to create the target observations (black crosses).}
	\label{fig:amortized}
\end{figure}

We chose a target parameter $\myvec{\theta}^*$ at the center of the parameter space to measure target observations $\myvec{x}^*$.
For the experiment with the two-dimensional parameter space and the \gls{psp} heights for an input to the first compartment, we want to show that the approximated posterior is also appropriate for other choices of the target parameter $\myvec{\theta}^*$.
As mentioned in the introduction, the posterior estimation is amortized after the first round of \gls{snpe} and can therefore be used to infer parameters $\myvec{\theta}$ for any observation $\myvec{x}$.
We draw five random parameters $\{\myvec{\theta}^*_i\}$ from the uniform prior and emulate the model on \gls{bss2} with the given parameters to record observations $\{\myvec{x}^*_i\}$.
For each of these observations, we draw samples from the amortized posterior estimation $\myvec{\theta} \sim p(\myvec{\theta} \mid \myvec{x}^*_i)$, \cref{fig:amortized}.

For each of the randomly selected observations $\myvec{x}^*_i$ the drawn samples cluster around the parameters which were used to obtain the given observation $\myvec{\theta}^*_i$.
Even if the target parameters are at the edge of the parameter space, the approximated posterior returns samples near these target parameters.
Therefore, we conclude that the \gls{snpe} algorithm is suitable to find parameters for observations which were obtained for parameters at arbitrary locations in the parameter space and that our choice of target parameters $\myvec{\theta}^*$ at the center of the parameter space does not affect the generality of the reported results.

\subsection{Simulations}

\begin{figure*}
	\includegraphics{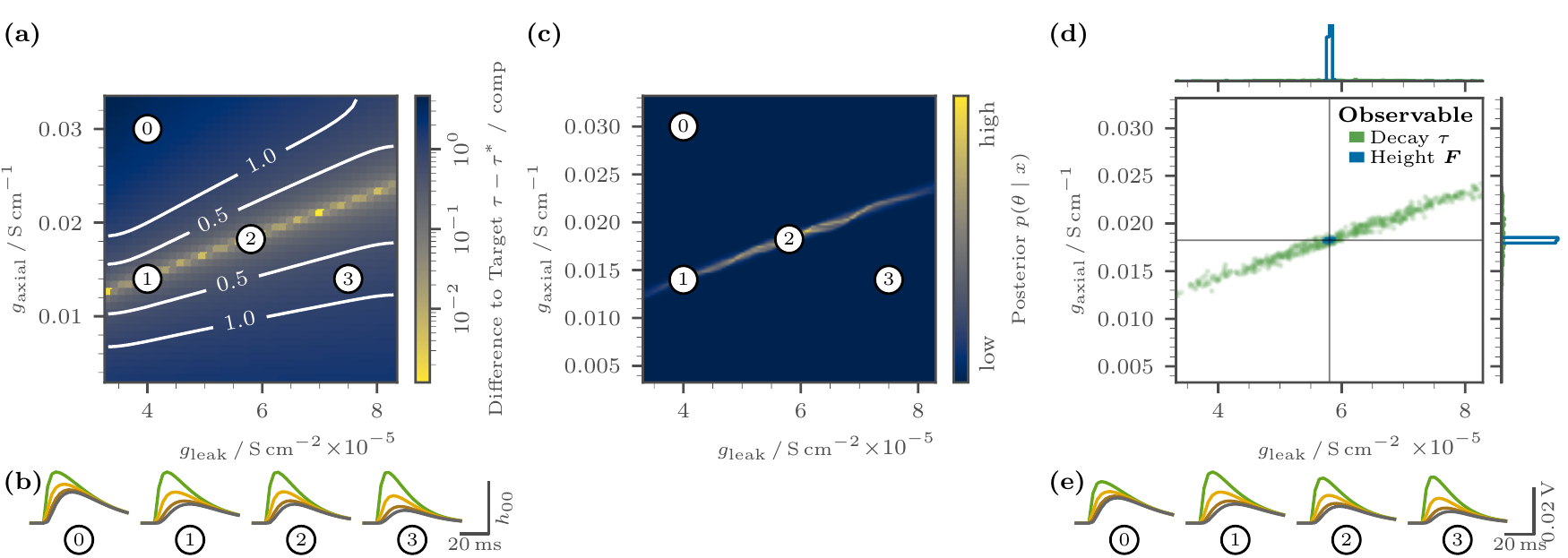}
	\caption{
		Propagation of \glsfmtfirstpl{psp} in a passive chain of four compartments simulated in Arbor.
		We performed the same experiments as in \cref{fig:2d} and we follow the structure of this figure.
		\subcaption{A} Grid search of the decay constant $\tau$.
			The dependency on the leak conductance $g_\text{leak}$ and the axial conductance $g_\text{axial}$ is comparable to \figGridsearchDiff{}.
		\subcaption{B} Example traces recorded at different locations in the parameter space, compare panel \formatpanel{A}.
			The traces are scaled relative to the height in the first compartment $h_{00}$.
		\subcaption{C} Posterior obtained with the \glsfmtfirst{snpe} algorithm.
			While the shape of the approximated posterior is comparable to the one in \figPosterior{}, the approximated posterior for the simulations is narrower.
		\subcaption{D} \num{500} random samples drawn from the approximated posteriors for two different types of observations.
		The distribution of the random samples is comparable to the results in \figSamples{}, but in agreement with the narrower posterior in panel \formatpanel{C}, the distribution of the samples is more narrow.
		\subcaption{E} Same traces as in panel \formatpanel{B} but shown on an absolute scale.
	}
	\label{fig:arbor}
\end{figure*}

We used the Arbor simulation library (version \texttt{0.8.1}) to compare our results to computer simulations \citep{abi2019arbor}.
Arbor is a high-performance simulator which supports multi-compartmental neuron models.
As Arbor solves the model equations numerically, it does not suffer from trial-to-trial variations and thus we expect the posterior distributions to be narrower.

As in the main part of the paper, we simulated a chain with four compartments.
The length of a single compartment was set to $l_\text{comp}=\SI{1}{\milli\meter}$, its diameter to $d_\text{comp}=\SI{4}{\micro\meter}$ and its capacitance to $C=\SI{125}{\pico\farad}$.
While the length and diameter were chosen arbitrarily, the capacitance reflected the capacitance of the compartments used during the emulation on \gls{bss2}.
The range of the leak conductance $g_\text{leak}$ was selected such that the membrane time constant of the simulated neurons was in agreement with the emulated neurons on \gls{bss2}.
Similarly, the range of the axial conductance $g_\text{axial}$ was chosen such that the axial conductance along a simulated compartment is comparable to the conductance between compartments on \gls{bss2}.

The results from the grid searches were comparable, \cref{fig:arbor} and \cref{fig:2d}, but the chosen parameter ranges led to a slightly higher dynamic range of the length constant.
In both cases a correlation between the leak and axial conductance was observed.

The shapes of the approximated posteriors also agreed with the results obtained for emulation on \gls{bss2}.
As expected, the approximated posterior distribution for the simulation was narrower than the approximation for \gls{bss2} due to temporal noise.
 
\end{document}